\documentclass{article}

\usepackage{microtype}
\usepackage{graphicx}
\usepackage{subfigure}
\usepackage[inline]{enumitem}
\usepackage{booktabs} 
\usepackage{nicefrac}

\usepackage{hyperref}

\usepackage[arxiv]{icml2024}

\usepackage{amsmath}
\usepackage{amssymb}
\usepackage{mathtools}
\usepackage{amsthm}

\DeclareMathOperator*{\argmax}{argmax}

\usepackage[capitalize,noabbrev]{cleveref}

\theoremstyle{plain}

\theoremstyle{definition}

\theoremstyle{remark}

\usepackage[disable]{todonotes}
\setuptodonotes{inline}
\usepackage{xargs}
\newcommandx{\katharina}[2][1=]{\todo[linecolor=cyan,backgroundcolor=cyan!25,bordercolor=cyan,#1]{#2}}
\newcommandx{\christoph}[2][1=]{\todo[linecolor=brown,backgroundcolor=brown!25,bordercolor=brown,#1]{#2}}
\newcommandx{\leon}[2][1=]{\todo[linecolor=violet,backgroundcolor=violet!25,bordercolor=violet,#1]{#2}}

\icmltitlerunning{Safe Active Learning for Gaussian Differential Equations}

\begin{document}

\twocolumn[
	\icmltitle{Safe Active Learning for Gaussian Differential Equations}

 	\icmlsetsymbol{equal}{*}

	\begin{icmlauthorlist}
		\icmlauthor{Leon Glass}{aaa}
		\icmlauthor{Katharina Ensinger}{aaa}
		\icmlauthor{Christoph Zimmer}{bbb}
	\end{icmlauthorlist}

	\icmlaffiliation{bbb}{Bosch Center for Artificial Intelligence, Robert Bosch GmbH, Renningen, Germany}
	\icmlaffiliation{aaa}{Robert Bosch GmbH, Renningen, Germany}

	\icmlcorrespondingauthor{Christoph Zimmer}{christoph.zimmer@de.bosch.com}
	\icmlkeywords{Machine Learning, ICML}

	\vskip 0.3in
]

\printAffiliationsAndNotice{}  

\begin{abstract}

	Gaussian Process differential equations (GPODE) have recently gained momentum due to their ability to capture dynamics behavior of systems and also represent uncertainty in predictions. Prior work has described the process of training the hyperparameters and, thereby, calibrating GPODE to data. How to design efficient algorithms to \textit{ collect data } for training GPODE models is still an open field of research.

	Nevertheless high-quality training data is key for model performance. Furthermore, data collection leads to time-cost and financial-cost and might in some areas even be safety critical to the system under test. Therefore, algorithms for safe and efficient data collection are central for building high quality GPODE models.

	Our novel Safe Active Learning (SAL) for GPODE algorithm addresses this challenge by suggesting a mechanism to propose efficient and non-safety-critical data to collect. SAL GPODE does so by sequentially suggesting new data, measuring it and updating the GPODE model with the new data. In this way, subsequent data points are iteratively suggested. The core of our SAL GPODE algorithm is a constrained optimization problem maximizing information of new data for GPODE model training constrained by the safety of the underlying system. We demonstrate our novel SAL GPODE’s superiority compared to a standard, non-active way of measuring new data on two relevant examples.

\end{abstract}

\section{Introduction}
Computational modeling is central in many domains from medicine~\cite{Winslow12} to epidemiology~\cite{brauer2019mathematical}, biology~\cite{mendes2009computational}, engineering~\cite{quarteroni2009mathematical} and even archaeology~\cite{schafer2011large}.
With the rise of machine learning the power of data based modeling has been recognized~\cite{jordan2015machine}.
More recently, hybrid modeling as a combination of first-principle-based domain knowledge and data-based components started playing a role~\cite{rackauckas2020universal,ott2023uncertainty,heinonen2018learning,yildiz2022learning,kurz2022hybrid}.

Models in general are calibrated to data.
This especially holds true for data-based and, therefore, also for data-based components in hybrid models.
This emphasizes the relevance of data collection.
Depending on the model and underlying system, data collection might be expensive.
This has been recognized early and there is a long history of efficient data collection from experimental design~\cite{atkinson1975optimal} to active learning~\cite{settles2009active} for classification and regression.
As data collection may be safety critical on some systems, safe learning methods have been developed~\cite{berkenkamp16bayesian,Schreiter15ecml,zimmer2018safe,li2022safe}.\\

Gaussian Processes (GP) are widely used in modeling~\cite{williams2006gaussian}, hybrid modeling~\cite{heinonen2018learning,yildiz2022learning} as well as active learning~\cite{Schreiter15ecml,zimmer2018safe} due to their inherent uncertainty quantification.

This paper proposes a novel algorithm for Safe Active Learning (SAL) in Gaussian Process ordinary differential equation (GPODE) models, named SAL GPODE.\@
Our novel SAL GPODE algorithm allows for safe and automated collection of highly informative data for dynamics systems of GPODE.\@

We suggest a new acquisition function to measure information content of possible new measurements in GPODE and derive a safety function evaluating the probability of a candidate for measurement to be safe.

The GOAL is to predict a dynamic model at any time point $t \in [0,T]$ with $T>0$.
This means that a standard GP is not applicable and neither are discrete-time state space models as they would need to be re-trained new for any $t'\neq t$.

Our contribution: A novel algorithm for Safe Active Learning in Gaussian Differential Equations including
\begin{itemize}
	\item a new acquisition function measuring information of new candidate measurements
	\item a new safety function allowing us to evaluate the safety of a new candidate measurement
	\item experimental results on a relevant benchmark showing our competitive performance
\end{itemize}

\begin{figure}\label{fig:enter-label}
	\centering
	\includegraphics[width=1\linewidth]{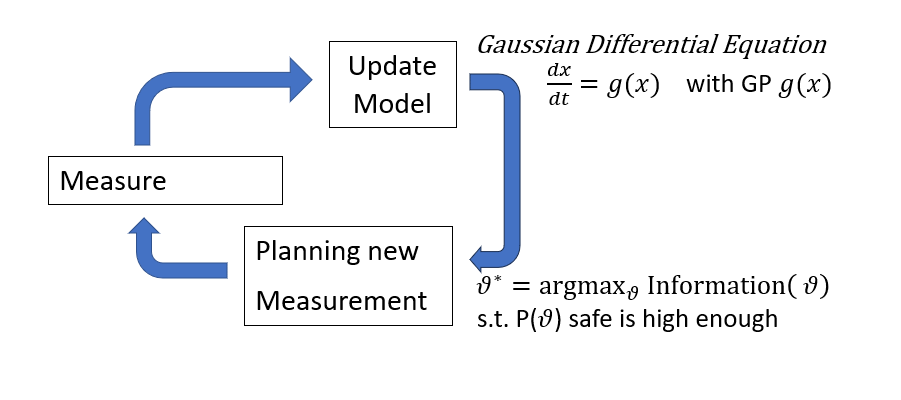}
	\caption{
		The visualization shows the Safe Active Learning loop, from upper right:
		i)  train/update a GPODE model,
		ii) plan new measurement by solving the constrained optimization problem that maximizes information constraint on safety,
		iii) conduct new measurement on the test system.
	} 
\end{figure}

\section{Problem statement}\label{sec:problem}
We focus on dynamic systems with $d$ dimensional states $x\in\mathbb{R}^d$ that evolve over time $t\in[0,T]$ with $T>0$ 
\begin{gather}
	\label{eq:gpode}
	\frac{dx}{dt} = f(x) ,
\end{gather}
where
$g:\mathbb{R}^d\mapsto\mathbb{R}^d$ represents an unknown component that will be modeled in a data-based fashion.

Noisy measurements $y_{1:N} = y_1,\ldots,y_N$ at $N$ time points $t_1,\ldots,t_N$ are available with $y_i = x(t_i) + \epsilon$ with $\epsilon\sim\mathcal{N}(0,\sigma)$ and $\sigma > 0$.
We introduce the GPODE time series modeling technique in Subsection~\ref{sec:GPODE}.

We sometimes use $y_{1:N}(\vartheta)$ with $\vartheta = x_0$ to emphasize the dependence of measurements on the active choice of an initial value $x_0$.

\textbf{Aim} We aim at finding $M$ new initial states $x_0^{(i)}$, $i=1,\ldots,M$ that lead to time series measurements $y_1^{(i)},\ldots,y_N^{(i)}$ that are most informative.

\textbf{Challenge}.
\begin{itemize}
	\item In contrast to most active learning problems we cannot measure the output $g(x)$ of our data-based model of interest, not even in a noisy fashion, but only the resulting time series $y_1,\ldots,y_N$.
	\item Even if we \textit{knew} which $g(x)$ would currently be most informative, we would not know for certain how to guide our system there as our systems model $x$ includes the uncertain data-based model $g(x)$.
\end{itemize}

\section{Related work:}
Active Learning~\cite{settles2009active} for classification~\cite{Joshi09,houlsby2011bayesian} or regression~\cite{yue2020active,garnett2013active,bitzer2022structural,yu2021active} chooses highly informative samples.
Safe Active Learning~\cite{Schreiter15ecml,zimmer2018safe,li2022safe,tebbe2024efficiently} takes additional safety constraints into account.
None of these works considers Gaussian Differential Equations or hybrid models.

Safe Bayesian Optimization~\cite{sui2018stagewise,berkenkamp2016safe} aims at finding an optimal configuration for a system and, therefore, has a different goal from Active Learning trying to build an overall good model (e.g., for acting as a digital twin or later worst-case analysis).

Recently, neural, Bayesian neural and Gaussian differential equations describe the right-hand side of a differential equation or part of it with a neural network or Gaussian Process~\cite{heinonen2018learning,yildiz2022learning,ott2023uncertainty,rackauckas2020universal,ensinger2024exact,dandekar2020bayesian}.
However, none of these works considers Active or Safe Active Learning.

Continuous time modeling techniques typically require the use of specialized ODE solvers to deal with stiff systems effectively~\cite{Rico-Martinez1993,Kim2021}.
Incorporating known physical structure may exacerbate the need for stiff solvers in such models~\cite{kidger2021neural}.
For this work, we do not consider stiff systems.
In principle, our method is composable with stiff ODE-solvers.
Further research is needed to investigate the interaction between stiff solvers and the performance of the proposed active learning scheme.

State space models~\cite{doerr2018probabilistic} and Active Learning for state space models~\cite{yu2021active} as well as~\cite{achterhold2021explore} and~\cite{buisson2020actively} focus on learning the transition from one time point to another.
This is from a modeling perspective different from universal differential equations that aim at learning the instantaneous transition.
Additionally, these methods do not consider hybrid models.

Physics informed neural networks are also a hybrid approach but to some extent orthogonal to universal differential equations as they focus on \textit{solving} an existing differential equation instead of \textit{enhancing} the solution as in UDE.\@
Therefore, also Active Learning for Physics informed neural networks~\cite{arthurs2021active,aikawa2024improving} is complementary.

There is one preprint~\cite{plate2024optimal} on experimental design for neural differential equations.
However, this piece of work does not address safe data generation or Gaussian Differential Equations.

\section{Background}
A \textbf{Gaussian Process} (GP) is a data based modeling technique that allows to model a mapping $g: \mathcal{X} \subseteq \mathbb{R}^d \rightarrow \mathbb{R}$ between inputs $x\in\mathcal{X}$ and outputs $ g(x) \in\mathbb{R}$ with $d\in\mathbb{N}$.
A GP is fully defined by its mean function 
and its kernel function $k: \mathcal{X}\times\mathcal{X}\rightarrow\mathbb{R}$ describing the similarity $k(x,x')$ between elements $x$ and $x'$ using hyperparameters $\theta$.
Given pairs of inputs $x_i$, $i=1,\ldots,N$ and possible noisy outputs $y_i = g(x_i) + \varepsilon_g$ with $\varepsilon_g \sim \mathcal{N}(0,\sigma_g)$ with $\sigma_g > 0$, the likelihood function $\mathcal{L}(D,\theta)$ measures how well the hyperparameters $\theta = (l, \sigma)$ with lengthscales $l=(l_1,\ldots,l_d)$ describe the data $D={(x_i,y_i)}_{i=1}^N$.
Hyperparameter training selects the optimal hyperparameters $ \theta^* = \argmax \mathcal{L}( D, \theta )$.
A posterior distribution can then predict the output of a new input $x^*$ as $g(x^*) \sim \mathcal{N}( \mu(x) , \sigma_{\varepsilon}(x) )$ with the GP's posterior mean $\mu$ and variance $\sigma_{\varepsilon}^2$.
The textbook~\cite{williams2006gaussian} provides a detailed description of GPs and their properties.
A major benefit of GPs leading to their wide usage is the natural uncertainty quantification of model predictions.\\[12pt]
\textbf{Variational Sparse GPs}.
Variational sparse GPs reduce GP complexity by approximating the posterior with a smaller amount of so-called inducing inputs and outputs that can be interpreted as pseudo observations~\cite{Titsias2009, Hensman2015}.
Analogous to~\cite{Hegde2021, doerr2018probabilistic}, we make use the formulation of~\cite{Hensman2015}.
\textbf{Decoupled Sampling} provides an elegant technique in order to sample complete functions from the GP posterior with linear complexity~\cite{wilson2020efficiently}.
The approach is especially beneficial for sampling trajectories from GP dynamics.
This is due to the fact that the dynamics function depends on all states previously visited in the current trajectory, leading to an iterative sampling process~\cite{Hewing2020}.
Thus, standard GP inference is computationally intractable for this task since each function draw has to be conditioned not only on the training data but also on all previous samples.
On a technical level, decoupled sampling splits the non-stationary posterior into the GP prior and an update via Matheron's rule.
In summary, for a sparse GP with $L$ inducing inputs this yields
\begin{equation}
	\begin{aligned}\label{eq:matheron}
		g(x^{\star}|Z,U) & =\underbrace{g(x^{\star})}_{\textrm{prior}}+\underbrace{k(x^{\star},Z)K^{-1}(U-g_Z)}_{\textrm{update}} \\
		                 & \approx \sum_{i=1}^{S} w_i \phi_i(x^{\star})+\sum_{j=1}^L v_j k(x^{\star},Z_j),
	\end{aligned}
\end{equation}
where $K \in \mathbb{R}^{L \times L}$ denotes the kernel matrix and $k(x^{\star},Z)$ the cross kernel between query point and inducing inputs.
The stationary GP prior is approximated with $S$ Fourier bases $\phi_i$ and $w_i \sim \mathcal{N}(0,1)$~\cite{Rahimi2008}.
For the update in Eq.~\eqref{eq:matheron}, it holds that $v = K^{-1}(U-\Phi w)$ with feature matrix $ \Phi = \phi(Z_i) \in \mathbb{R}^{L \times S}$ and weight vector $w \in \mathbb{R}^{S \time 1}$.
In this work, we leverage decoupled sampling in order to draw complete dynamics functions from the GP posterior for both, training and active learning.

\textbf{Gaussian Differential Equations}\label{sec:GPODE}
In this work, we leverage the modeling technique from~\cite{Hegde2021}.
They model the dynamics $g$ of an ODE $\dot{x}(t)=g(x(t))$ with a zero mean GP, where noisy observations $y_1,\dots,y_N$ are given.
For $L$ GP inputs $Z \in \mathbb{R}^{L \times d}$ and outputs $U=g(Z) \in \mathbb{R}^d$, this yields the factorization
\begin{equation}
	p(y_{1:N},g,U,x_0)\sim \prod_{n=1}^N \mathcal{N}(y_{n}|x_{n},\sigma^{2})p(g,U)p(x_0),
\end{equation}
where $p(g,U)=p(g|U)p(U)$ denotes the joint probability of a GP sample and the inducing outputs.
Further, $p(U)=\mathcal{N}(0,K)$, where $K \in \mathbb{R}^{L \times L}$ denotes the kernel matrix evaluated at the inducing inputs.
The states are obtained by solving the ODE, thus $x_i=x(t_i)=x_0+\int_0^{t_i}g(x(s))ds$ with initial value $x_0$.
Since $U,Z$ and $x_0$ are not given during training, the posterior $p(U,g,x_0|y_{1:N})$ is approximated via a variational posterior
\begin{equation}
	q(U,g,x_0) \sim p(g|U)q(U)p(x_0),
\end{equation}
where $q(U)\sim \mathcal{N}(\mu_u,\Sigma_u)$.

In contrast to \cite{Hegde2021}, we do not learn a variational distribution for the initial state $x_0$.
Instead, we choose a normal distribution centered around the first state measurement with a fixed standard deviation, which works robustly in our experiments.
Note that in Active Learning, we \textit{choose} $x_0$ in contrast to the standard modeling task.
Still, we do not include information about the noise-free initial value during model training in order to allow consistency with the modeling framework.

During training and prediction, consistent dynamics $g$ are obtained via decoupled sampling~\cite{wilson2020efficiently}.
The parameters $\mu_u$, $\Sigma_u$ and GP hyperparameters are adapted by maximizing the evidence lower bound (ELBo).
Similar to neural ODEs, the GP dynamics are integrated with an adaptive step size integrator during training and prediction.

\textbf{Active Learning} aims at sequentially selecting informative samples $x^*$ based on an acquisition function $\alpha$ as a metric for information content:
\begin{gather*}
	x^* = \argmax_{x\in \mathbb{X}} \alpha( x)
\end{gather*}
Various candidates for acquisition function have been suggested in machine learning such as measures of disagreement~\cite{settles2009active}, mutual information or entropy~\cite{krause2008near}.
In case of GP models, most prominently, entropy and mutual information are used.
Entropy has the special advantage of being proportional to predictive variance in case of Gaussian distribution and, therefore, commonly $\alpha(x) = \sigma(x)$ which can be calculated in closed form.

\textbf{Safe Active Learning} is employed if the technical system that data is queried from can exhibit safety critical behavior for certain inputs $x$.
Some of the safety critical behavior can be prevented by avoiding e.g.
too high or too low values of $x$.
This can easily be implemented by using constraints on the input space $\mathbb{X}$.

The more difficult case that we consider here is if there is a certain threshold, e.g., $s_{crit}$ in the output space of a safety related quantity $z$.
To address this, we learn an additional GP, the safety model $s:\mathbb{X}\rightarrow\mathcal{Z}$ with a safety related quantity $z\in\mathcal{Z}$ that should stay below a threshold $s_{crit}$.
Now, we can calculate the probability $\xi(x)$ that the safety output $z$ of the considered input $x$ is below the critical threshold $s_{crit}$:
\begin{gather*}
	\xi(x) = P( z \leq s_{crit} ) = \int_{ - \infty}^{z_{crit}}\mathcal{N}(z|\mu_s(x),\sigma_s^{2}(x) )
\end{gather*}
with $\mu_s$ and $\sigma_s$ begin mean and standard deviation of the safety GP $s$.

The active learning problem is then modified to
\begin{subequations}
	\begin{alignat}{2}
		x^* =\, & \argmax_{x\in \mathbb{X}} & \quad & \alpha( x)                                \\
		        & \text{s. t.}              &       & \xi(x)  \geq                       \delta 
	\end{alignat}
\end{subequations}
with some small user-chosen $0 \leq \delta < 1$ that may depend on whether the safety violation just causes some time delay (smaller $\delta$ thinkable) or critically damages the technical system (higher $\delta$).

\section{Safe Active Learning for Time-Series Modeling}\label{Sec:DSAL}

We refer to our problem statement for dynamics modeling with GPODE and its related challenges in efficient data collection given in Section~\ref{sec:problem}.

\subsection{How to measure information in GPODE?}\label{subsec:al}
We focus on the mutual information between a time series measurement $y_{1:N}(\vartheta)$ and our data based model of interest $g$:
\begin{equation}
	\label{eq:mi}
	I(y_{1:N}(\vartheta),g) \vcentcolon = H(y_{1:N}(\vartheta))-H(y_{1:N}(\vartheta)|g),
\end{equation}
where we use the entropy-based definition for the mutual information of the infinite-dimensional Gaussian Process $g$ and the observation $y_{1:N}$.
Conditioned on the GP $g$, the observation $y_{1:N}$ is determined up to measurement noise $\sigma$.
Therefore, the second term does not depend on an initial value
and can be neglected in the later information maximization problem.

The marginal entropy is
\begin{equation}
	\begin{aligned}
		H(y_{1:N}) & =-\int p(y_{1:N})\log(p(y_{1:N}))                                                                       \\
		           & =-\mathbb{E}_{y_{1:N} \sim p(y_{1:N}) }\log(p(y_{1:N}))                                                 \\
		           & =-\mathbb{E}_{y_{1:N} \sim p(y_{1:N}|g)p(g) } \log \int p(y_{1:N}|g)p(g) dg                             \\
		           & =-\mathbb{E}_{y_{1:N} \sim p(y_{1:N}|g) p(g))} \log \left(\mathbb{E}_{g \sim p(g)} p(y_{1:N}|g)\right). 
	\end{aligned}
\end{equation}
We draw  $K$ samples $g^{1:K}\sim p(g)$ from the Gaussian process by decoupled sampling (see Background subsection Decoupled Sampling).
We compute $x_{1:N}^l$ from $g^l$ by numerical integration with equation~\eqref{eq:gpode}.
Similarly, we calculate $y_{1:N}^{1:K}\sim p(y_{1:N}|g)$ by additionally adding noise.

With that, we can approximate the entropy with Monte-Carlo sampling:
\begin{equation} \label{eq:entropy}
	H(y_{1:N}) \approx -\frac{1}{K} \sum_{m=1}^K \log\left(\frac{1}{K}\sum_{l=1}^K \mathcal{N}(y_{1:N}^m|x_{1:N}^l,\sigma^2)\right)
\end{equation}

The acquisition function for the \textit{active} learning is, hence,
\begin{equation}
	\label{eq:acq}
	\alpha(\vartheta) = I(y_{1:N}(\vartheta), g) \propto H( y_{1:N}(\vartheta)).
\end{equation}

We also suggest a \textbf{covariance-based acquisition function} as alternative.
The alternative focuses on the uncertainty of the sampled trajectories $x_{1:N}^{1:K}$ measured by its empirical covariance $Cov_{emp}$:
\begin{equation*}
	\label{eq:acq-alt}
	\alpha_2(\vartheta) = Cov_{emp}( x_{1:N}^{1:K}(\vartheta) )
\end{equation*}

Note that the second acquisition function $(\ref{eq:acq-alt})$ is equivalent to the first $(\ref{eq:acq})$  in case of $x_{1:N}^{1:K}$ following a multi-variate normal distribution.
However, this will in general not be the case with differential equation based trajectories, so the covariance-based acquisition function is different.

\subsection{How can we check whether a new measurement is safe?}
We have introduced the concept of safety in safe learning problems as a constrained in the output space that should be met.
As we act in a state space setting here, constraints will be phrased in terms of the state space $\mathbb{X}$: $x^{\min} = x_1^{\min},\ldots,x_d^{\min}$ and $x^{\max}=(x_1^{\max},\ldots,x_d^{\max})$.
Inactive dimensions can be treated, e.g., as $x_i^{\max} = \infty $.
As usual, safety can have a wide range of meanings from really critically to the system's health to rather an inconvenience due to time loss in case of violation.

We are interested in the probability of a safety violation given a possible choice of an initial value: $P( \vartheta \text{ leading to unsafe behavior} )$.

A behavior is unsafe if for any $t$ the constraints are violated.
As in~\cite{zimmer2018safe} we resort to a discretization, hence
\begin{gather}
	\label{eq:safe}
	\xi(\vartheta) =  P( x^{\min} \leq x(t_i) \leq x^{\max}\ \forall i ) \geq \delta
\end{gather}
with a user defined probability $0 \leq \delta < 1$.
The inequality is meant to hold for all dimensions of $x$.

As in the Subsection~\ref{subsec:al}, we draw  $K$ samples $g^{1:K}\sim p(g)$ from the Gaussian process by decoupled sampling and calculate $x^{1:K}(t_i), i=1,\ldots,N$ by integration.

\subsection{Planning new GPODE trajectories with Safe Active Learning}
Now, that we have defined our measure of information content~\eqref{eq:mi} and our safety criterion~\eqref{eq:safe}, we can state the constrained optimization problem to determine the choice of an initial value $\vartheta^*$ for the next safe and information optimal measurement:
\begin{subequations}\label{eq:constr-opt}
	\begin{alignat}{2}
		\vartheta^* =\, & \argmax_{\vartheta\in\Theta} & \quad & \alpha(y_{1:N}(\vartheta);g) \\
		                & \text{s. t.}                 &       & \xi(\vartheta) \geq \delta   
	\end{alignat}
\end{subequations}

\section{Experiments} 

\subsection{Van der Pol Oscillator}
Our first test case is the Van der Pol Oscillator as in~\cite{Hegde2021}.
Its dynamics are given by
\begin{subequations}
	\begin{align}
		\frac{dx_1}{dt} & = x_2                      \\
		\frac{dx_2}{dt} & = \mu{(1-x_1)}^2\ x_2-x_1.
	\end{align}
\end{subequations}
For our experiments, we generate data with $\mu=\nicefrac{1}{2}$.
We assume the complete right-hand side to be unknown and learn it efficiently by sequentially choosing $M$ initial states $x_0^{(i)} = {(x_1(0), x_2(0))}^{(i)}, i=1,\ldots,M$.
We impose the safety constraint
\begin{equation}
	-4 \leq x_{i}(t) \leq 4,\quad \forall t \in \left[0,3\right], i=1,2.
\end{equation}
Active learning starts from an initial dataset created by drawing a single initial state $(x_{1}(0), x_{2}(0)) = (-1.5, 2.5)$.

\begin{figure}[H]
	\includegraphics[width=1\linewidth]{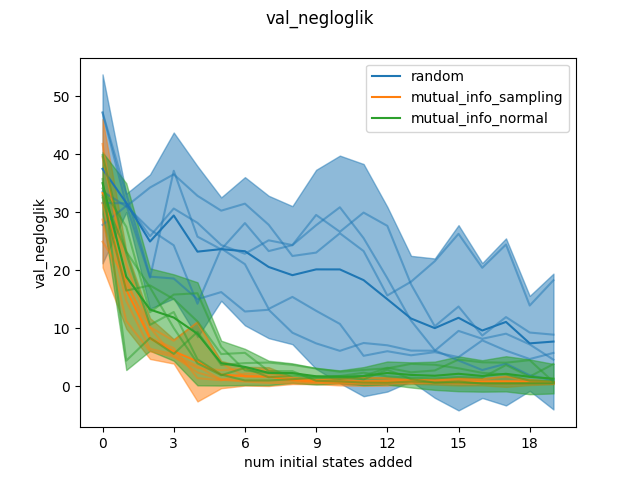}
	\caption{SAL GPODE outperforming random benchmark: Validation negative log likelihood as a performance indicator in dependence of number of measurements. Lower values mean smaller error. Our SAL GPODE quickly reduces the error and as already learned a good model with as few as 2--3 measurements. The random benchmark needs much longer. Solid line is the mean and shaded area the mean plus/minus two standard deviations of five repetitions with different random seeds.}
    \label{fig:vdp-sal-bench}
\end{figure}

Figure~\ref{fig:vdp-sal-bench} shows that our novel SAL GPODE algorithm learns the systems dynamic much quicker than the random baseline.

\begin{figure}[h]
	\centering
	\includegraphics[width=\linewidth]{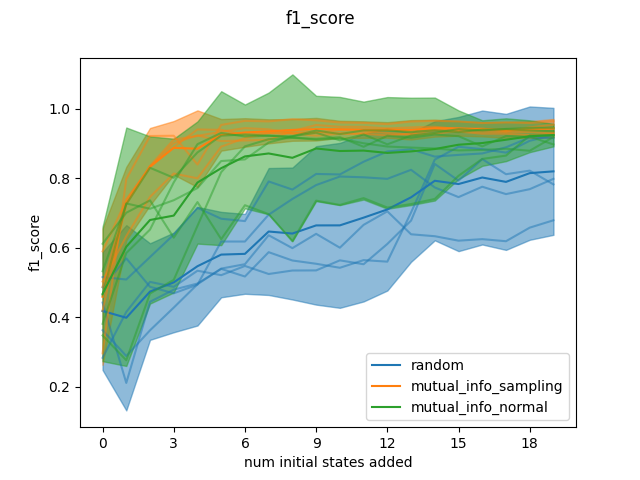}
	\caption{Exploration of the safe set}
    \label{fig:vdp-f1}
\end{figure}

Figure~\ref{fig:vdp-f1} shows that our novel SAL GPODE algorithm is also able to learn the safe area more quickly. We check how many initial states are correctly identified as safe or unsafe with our SAL GPODE in Equation (\ref{eq:safe}) and how many are incorrectly identified and display the F1 score. 

\subsection{Lotka-Volterra}
The Lotka-Volterra model describes the population dynamics of an interacting pair of predator and prey species.
Its dynamics are given by
\begin{subequations}
	\begin{align}
		\frac{dx_{1}}{dt} & = \alpha x_{1} - \beta x_{1}x_{2}    \\
		\frac{dx_{2}}{dt} & = -\gamma x_{2} + \delta x_{1}x_{2},
	\end{align}
\end{subequations}
where $x_{1}$ is the prey population, $x_{2}$ is the predator population and $\alpha, \beta, \gamma, \delta > 0$ are fixed parameters.
We choose the parametrization
\begin{equation}
	\begin{aligned}
		\alpha & = 0.5 & \beta  & = 0.05 \\
		\gamma & = 0.5 & \delta & = 0.05
	\end{aligned}
\end{equation}
for our simulator.

\begin{figure}[h]
	\centering
	\includegraphics[width=\linewidth]{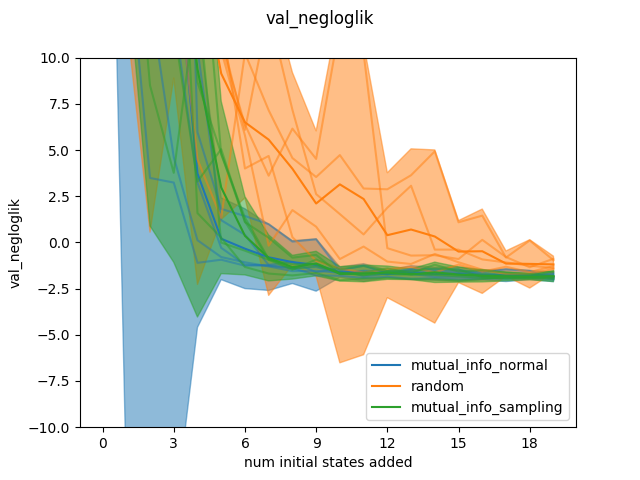}
	\caption{Validation negative log likelihood on the Lotka-Volterra task}
    \label{fig:lotka-volterra-negloglik}
\end{figure}

\begin{figure}[h]
	\centering
	\includegraphics[width=\linewidth]{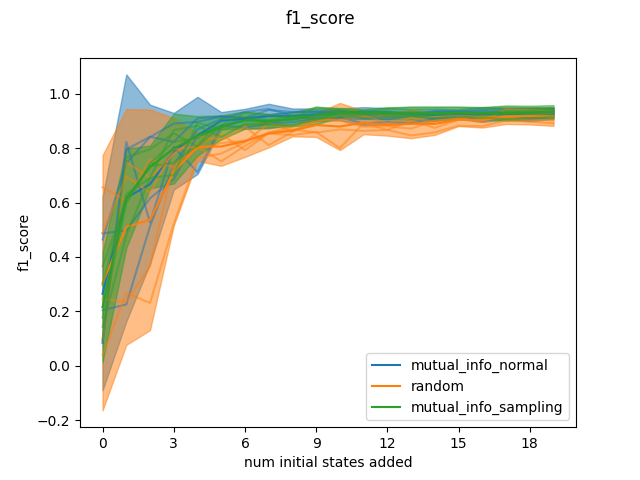}
	\caption{F1 score on the Lotka-Volterra task}
    \label{fig:lotka-volterra-f1}
\end{figure}

Figure~\ref{fig:lotka-volterra-negloglik}~and~\ref{fig:lotka-volterra-f1} demonstrate the superiority of our approach as the negative log-likelihood drops faster and the F1 score increases faster than with the non-active benchmark.

\section{Limitations and Future work}

Hyperparameter estimation (Subsection~\ref{sec:GPODE}) requires a numerical optimization and is, therefore, not free of computational cost.
Recently, amortization techniques have been developed in the area of Gaussian Processes~\cite{liu2020task,bitzer2023amortized} to address this by, e.g.,\ upfront training a neural network based transformer with lots of synthetic data sets that allows to predict hyperparameters for a new data set by just a forward evaluation of the neural network.
Extending this work to GPODE is orthogonal to our idea but could, nevertheless, be beneficial for future work.

This work uses a fixed kernel for the GP.\@ Depending on the problem and specific knowledge e.g.\ on the structure, one could extend the work on kernel selection~\cite{bitzer2022structural} to the scenarios of GPODE.\@

While many phenomenons can be described with ODEs, future work could also address systems that require partial or stochastic differential equations for modeling.

\section{Summary}
We proposed SAL GPODE, a novel algorithm for automated collection of safe and informative data for GPODE.\@
Our experiments show that SAL GPODE outperforms conventional data collection schemes in terms of model learning efficiency.

\section*{Acknowledgement}
This work is funded by the German Federal Ministry for Economic Affairs and Climate Action as part of the project “KI-Grundlagenentwicklung mit Leitanwendungen Virtuelle Sensorik und Brennstoffzellenregelung (KI-Embedded)” (FKZ: 19I21043A). The financial support is hereby gratefully acknowledged.

\bibliography{AL-GPODE-01}
\bibliographystyle{icml2024}

\newpage
\appendix
\onecolumn

\section{Conditional entropy}
The conditional entropy is defined as
\begin{equation}
	\mathbb{E}_{f \sim p(f)}(-\log\left(p(y_{1:N}|f))\right).
\end{equation}
We approximate the expectation in a sampling-based manner again.
This yields
\begin{equation}
	H(y_{1:N}|f) \approx \frac{1}{M} \sum_{m=1}^M \log p(y_{1:N}^m|x_{1:N}^m).
\end{equation}
Since the observations are sampled by adding noise to the trajectories this is a constant, and we can ignore this part.

\end{document}